\pdfoutput=1

\documentclass[11pt]{article}

\usepackage[preprint]{acl}

\usepackage{times}
\usepackage{latexsym}

\usepackage[T1]{fontenc}

\usepackage[utf8]{inputenc}

\usepackage{microtype}

\usepackage{inconsolata}

\usepackage{graphicx}

\usepackage{float} 
\usepackage{graphicx} 
\usepackage{multirow} 
\usepackage{geometry} 
\usepackage{caption} 

\usepackage{caption}
\usepackage[separate-uncertainty=true]{siunitx}
\usepackage{mathtools}
\usepackage{graphicx}
\usepackage{courier}
\usepackage{epstopdf}
\usepackage{color}
\usepackage{csquotes}

\usepackage{amsmath}
\usepackage{amsfonts}
\usepackage{amssymb}
\usepackage[subnum]{cases}
\usepackage{booktabs} 
\usepackage{array}
\usepackage{pifont}
\usepackage{colortbl,array,xcolor}
\usepackage[flushleft]{threeparttable}
\usepackage{arydshln}
\usepackage{float}

\usepackage{amssymb}
\usepackage{pifont}

\usepackage {tikz}
\usetikzlibrary {positioning}
\definecolor {processblue}{cmyk}{0.96,0,0,0}
\usepackage{lipsum,adjustbox}

\usepackage[subnum]{cases}
\usepackage{color,soul}

\usepackage{subcaption}

\usepackage{footnote}
\makesavenoteenv{tabular} 

\usepackage{multirow}
\usepackage{booktabs,subcaption,amsfonts,dcolumn}

\usepackage{xcolor}

\usepackage{textcomp}

\usepackage{bbold}
\usepackage{mathtools}
\usepackage{breqn}
\usepackage{tabularray}
\usepackage{amsthm}

\usepackage{makecell}

\usepackage[many]{tcolorbox}
\newtcolorbox{boxA}{
    fontupper = \bf,
    boxrule = 1.5pt,
    colframe = blue 
}

\newcommand{\method}{\emph{ID-SPAM}}
\definecolor{Gray}{gray}{0.90}
\newcommand\cg{\cellcolor{Gray}}

\title{Leveraging Self-Attention for Input-Dependent Soft Prompting in LLMs}

\author{
  Ananth Muppidi\thanks{Equal contribution. Work done during the internship at Adobe Research.} \\
  IIIT Hyderabad \\
  India \\
  {\small\texttt{ananth.muppidi21@gmail.com}}
  \And
  Abhilash Nandy\footnotemark[1] \\
  IIT Kharagpur \\
  India\\
  {\small\texttt{nandyabhilash@gmail.com}}
  \And
  Sambaran Bandyopadhyay \\
  Adobe Research \\
  India \\
  {\small\texttt{samb.bandyo@gmail.com}} 
}

\begin{document}
\maketitle
\begin{abstract}
The performance of large language models in domain-specific tasks necessitates fine-tuning, which is computationally expensive and technically challenging. This paper focuses on parameter-efficient fine-tuning using soft prompting, a promising approach that adapts pre-trained models to downstream tasks by learning a small set of parameters. We propose a novel Input Dependent Soft Prompting technique with a self-Attention Mechanism (\method) that generates soft prompts based on the input tokens and attends different tokens with varying importance. Our method is simple and efficient, keeping the number of trainable parameters small. We show the merits of the proposed approach compared to state-of-the-art techniques on various tasks and show the improved zero shot domain transfer capability.
\end{abstract}

\section{Introduction}\label{sec:intro}

\begin{figure*}[t]
    \centering
    \includegraphics[width=0.74\textwidth]{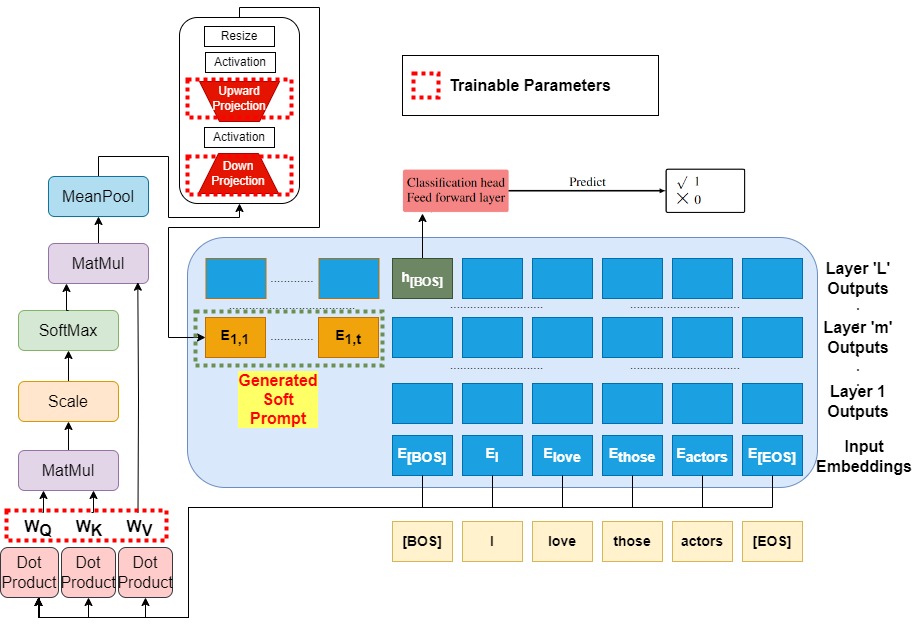}
    \caption{\method\ Framework. Given an LM, the generated soft-prompt can be prepended to any transformer layer's inputs (the figure can be best seen in color)}
    \label{fig:id-spam}
\end{figure*}

Large language models (LLMs) have made significant advancements in natural language processing tasks, such as generation, translation and summarization \citep{yeo2023assessing,zhang2023summit}. Despite their success, LLMs' performance in domain-specific tasks is limited, and fine-tuning on task-oriented datasets is crucial. As models from BERT \cite{devlin2019bert} to GPT-3 \cite{gpt3} have millions to billions of parameters, fine-tuning becomes computationally expensive and challenging. Therefore, parameter efficient fine-tuning \citep{han2024parameter} research aims to adapt pre-trained models to downstream tasks by fixing most parameters and only learning a small subset.

Soft prompting is a promising direction for fine-tuning large models. Without changing the core architecture of an LLM, soft prompt methods generally introduce a small trainable vector (known as a `soft prompt') at the beginning of one or more transformer layers' inputs within the LLM. During fine tuning, only the soft prompt is trained to adapt to the downstream task keeping the parameters of the base LLM frozen.
\citet{lester2021power} propose Prompt Tuning by prepending the trainable soft prompt vector before the embeddings of the text input, just after the embedding layer of the base LLM. On similar lines, \citet{li2021prefix} introduce Prefix Tuning by prepending a soft prompt at every transformer layer and \citet{liu2021gpt} come up with P-tuning by interleaving learnable prompts with input embeddings.
Contrary to text prompt engineering \cite{wei2022chain} or optimizing discrete token representations via in-context learning \citep{dai2023can}, \citet{petrov2023prompting} suggest that the continuous embedding space of soft prompts inherently possesses a greater amount of information. 

Recent literature introduces several variants of soft prompt techniques such as removing the reparameterization module \citep{liu2022p}, hierarchical structured pruning \citep{ma2022xprompt}, introducing an adaptive gate mechanism to control the prefix importance in each transformer layer \citep{zhang2023towards}, diving the soft prompt into query, key and value prompts \citep{wang2023aprompt}, learning multiple short soft prompts and a gating mechanism to route an input to a specific soft prompt \cite{choi2023smop}, and decomposing the soft prompt into low rank matrices \cite{shi2024dept}.

Many of these methods keep the soft prompt independent of the actual input given to the LLM. However, this limits the soft prompt to adjust based on the actual input during the inference time. It is unlikely that a unified prompt would lead to a performance improvement across different input instances. It also makes the training difficult by increasing the convergence time. To address this, a few recent approaches leverage input dependent soft prompts. But they need to concatenate the soft prompts either at every transformer layer of the base LLM \citep{wu2022idpg} or all the layers after an intermediate layer \citep{liu2022late}, or transform the soft prompt by using cross-attention with the input tokens without explicitly generating from them \citep{jin2023instance}.
These input dependent prompting techniques still have multiple limitations: (i) Many of them employ relatively complicated architecture by concatenating soft prompts in multiple internal transformer layers of the LLM; (ii) Since, a task may contain diverse samples with different types of words, it is important to attend different words of the input with different weights while generating the soft prompt; And (iii) Number of trainable parameters often increases significantly.

To address the above research gaps, we introduce an input dependent soft prompt technique where the soft prompt is generated by a trainable network that attends different tokens of the input with different importance by employing a self-attention mechanism. We prepend the soft prompt with the input to a single transformer layer of the base LLM, keeping the number of trainable parameters small and training smooth. Following are the \textit{contributions} made in this work: \textbf{(i)} We propose \method, a novel (\underline{I}nput \underline{D}ependent \underline{S}oft \underline{P}rompting technique with a self-\underline{A}ttention \underline{M}echanism); Our method is simple and efficient to train. \textbf{(ii)} We show the merit of the proposed approach on six tasks from the GLUE benchmark \citep{glue}; And \textbf{(iii)} Due to the use of trainable attention on the input tokens, our approach is more efficient in zero-shot domain transfer as shown in the experiment.

\section{Proposed Solution}\label{sec:soln}
In this section, we introduce our proposed method \method\ (see its framework in Figure \ref{fig:id-spam}). 


Given a Task $T$ having training data represented as $D_{train} = \{(x_i, y_i)\}_{i=1}^K$. Following \citet{lester2021power}, we represent the input as $x_i = \mathbf{E}(\texttt{[SEP]}S_1\texttt{[SEP]} S_2\texttt{[EOS]})$ for a task with a pair of sentences $S_1, S_2$ as the input or $x_i = \mathbf{E}(\texttt{[SEP]}S_1\texttt{[EOS]})$ for for a task with a single sentence $S_1$ as the input, where $\mathbf{E}(\cdot)$ is the token embedding for the input sentence(s). 

We introduce a learnable soft prompt such that the prompt not only varies with the task at hand, but is also generated based on the input in such a way that it primarily attends to those input tokens that are essential for the given task. To make the learning efficient, we freeze the parameters of the original LM $M$. Our proposed soft prompt for the task T can be defined as $\mathbf{S}_T\in\mathbb{R}^{n\times t}$, where $t$ is the number of tokens in the prompt representation and $n$ is the hidden dimension of the LM $\textbf{M}$ under consideration. $\mathbf{S}_T$ is obtained by first applying a \textit{learnable} attention layer \cite{attentionisallyouneed} over the input embeddings $\mathbf{E}(\cdot)$ and averaging the outputs, providing a context-rich representation. The $n \times 1$ dimensional vector $A$ so obtained is passed through a downward projection MLP Layer having learnable weights $\mathbf{W}_{down}\in\mathbb{R}^{n\times c}$ and bias $\mathbf{b}_{down}\in\mathbb{R}^{c}$, followed by a ReLU Activation Layer \cite{relu}, and then an upward projection MLP Layer having learnable weights $\mathbf{W}_{up}\in\mathbb{R}^{c\times n.t}$ and bias $\mathbf{b}_{down}\in\mathbb{R}^{n.t}$, where $c < n$. The output so obtained is re-sized to get the learnable, input-dependent soft prompt $\mathbf{S}_T\in\mathbb{R}^{n\times t}$, which is either prepended to the token embeddings or to the input of any intermediate transformer layer of the LM $\textbf{M}$. We will show some analysis on the choice of intermediate layer in the experiments. Mathematically, 
\begin{equation}
\small
 A = \text{mean}\Bigg{\{}\text{softmax}\left(\frac{\left(\mathbf{E} \mathbf{W}_Q\right)\left(\mathbf{E} \mathbf{W}_K\right)^\top}{\sqrt{d_k}}\right) \left(\mathbf{E} \mathbf{W}_V\right)\Bigg{\}}
\end{equation}
\begin{equation}
    \mathbf{S}_T = \text{resize}(\sigma(W_{up}\sigma(W_{down}(A))))
\end{equation}
$W_Q$, $W_K$, and $W_V$ are the query, key, and value parameter matrices respectively, and $\frac{1}{\sqrt{d_k}}$ is a scaling factor, as used in \citet{attentionisallyouneed}. $\sigma$ is a non-linear activation which we used ReLU here.

\section{Experimental Evaluation}\label{sec:exp}
Here, we describe our experimental setup, evaluate  \method\ framework on GLUE and SuperGLUE benchmarks, and zero-shot domain transfer between tasks against several baselines, followed by a detailed analysis.

\subsection{Experimental Setup}

We compare \method\ with the following baselines - (1) \textbf{Transformer fine-tuning:} Here, all parameters of LM are learned (2) \textbf{Parameter-Efficient Soft Prompt-based Methods -} (a) \textbf{Prompt Tuning:} We use standard prompt tuning \cite{lester2021power}, which learns soft prompts through backpropagation to condition frozen language models for specific tasks. (b) \textbf{P-tuning:} P-tuning \cite{liu2022p} is a variant of Deep Prompt Tuning \cite{li2021prefix, qin-eisner-2021-learning} adapted for NLU (c) \textbf{Sparse Mixture of Prompts (SMoP):} SMoP \cite{choi2023smop} leverages multiple short soft prompts with a gating mechanism to train multiple prompts tailored in addressing different data subsets (d) \textbf{Late Prompt Tuning (LPT):} LPT \cite{liu2022late} injects a late prompt into an intermediate layer of the LM, rather than into the input layer or across all layers. (e) \textbf{Decomposed Prompt Tuning (\textsc{DePT}): }\textsc{DePT} \cite{shi2024dept} employs a decomposition strategy for the soft prompt, breaking it down into a pair of low-rank matrices. These components are then optimized independently, each with its own specific learning rate. (3) \textbf{Parameter Efficient Fine-tuning using Low-Rank Adaptation (LoRA):} LoRA \cite{hu2022lora} addresses challenge of fine-tuning large language models by freezing pre-trained model's weights and introducing trainable low-rank matrices into each layer. Note that it does not use a soft prompt. 


For all methods, we train upto 30 epochs (\textit{\textbf{Section \ref{appendix:lora-convergence} of Appendix}} shows convergence after 30 epochs) using  Standard Cross-Entropy Loss and Adam Optimizer \cite{adamw}, and number of soft-prompt tokens $t = 10$. We perform hyperparameter tuning for \method, as described \textit{\textbf{in Section \ref{appendix:hyperparams} of Appendix}}. We use a NVIDIA A100 GPU with a VRAM of 80 GB for all experiments.

\subsection{Evaluation on GLUE Benchmark}

We evaluate \method\ and baselines on the following 6 Natural Language Understanding (NLU) Tasks from GLUE Benchmark \cite{glue} - SST-2 \cite{socher2013recursive}, MRPC \cite{dolan2005automatically}, MNLI \cite{williams2018broad}, QNLI \cite{rajpurkar2016squad}, RTE \cite{dagan2005pascal}, and QQP \cite{quora2017quora}. These tasks cover various aspects of natural language understanding and inference, providing a comprehensive assessment of our approach's performance across different language processing challenges. All datasets were obtained from the Hugging Face library \cite{wolf-etal-2020-transformers, lhoest-etal-2021-datasets}. Further dataset statistics are shared in Table \ref{table:dataset_statistics}.

\begin{table*}[htbp]
\centering
\resizebox{\textwidth}{!}{
\begin{tabular}{|c|c|c|c|c|c|c|c|}
\hline
\textbf{Category} & \textbf{Datasets} & \textbf{|Train|} & \textbf{|Dev|} & \textbf{|Labels|} & \textbf{Type} & \textbf{Labels} \\
\hline
Single-sentence & SST-2 & 67349 & 872 & 2 & sentiment & positive, negative \\
\hline
\multirow{5}{*}{Sentence-pair} & MNLI & 392702 & 19647 & 3 & NLI & entailment, neutral, contradiction \\
\cline{2-7}
& MRPC & 3668 & 408 & 2 & paraphrase & equivalent, not equivalent \\
\cline{2-7}
& QNLI & 104743 & 5463 & 2 & NLI & entailment, not entailment \\
\cline{2-7}
& QQP & 363846 & 40430 & 2 & paraphrase & equivalent, not equivalent \\
\cline{2-7}
& RTE & 2490 & 277 & 2 & NLI & entailment, not entailment \\
\hline
\end{tabular}
}
\caption{Statistics of the datasets used in our experiments.}
\label{table:dataset_statistics}
\end{table*}

We report accuracy for \{SST, MNLI, QNLI, RTE\} and average of accuracy and macro F1-Score for \{MRPC, QQP\} using RoBERTa-BASE, RoBERTa-LARGE backbones \cite{liu2019roberta} in Table \ref{table:main_results_glue_rbase}.

\begin{table}[t]
\centering

\resizebox{0.5\textwidth}{!}{
\begin{tabular}{lrccccccccc}
\toprule                                               
                 & MNLI & QNLI & SST-2  & MRPC  & RTE  & QQP & \cg Mean \\  \midrule
{\bf Method} &   \multicolumn{6}{c}{
\textbf{GLUE (RoBERTa-BASE Backbone)}} \\             \midrule     
Fine-tuning  & 87.4$_{2.4}$ & 91.3$_{1.0}$ & 92.3$_{0.6}$   & 92.7$_{0.7}$  & 82.5$_{1.3}$ & 90.9$_{0.8}$ & \cg 89.5 \\  
\hdashline\noalign{\vskip 0.4ex}
LoRA & 88.7$_{0.4}$ & 84.2$_{2.1}$ & 90.4$_{0.3}$ & 79.3$_{0.5}$ & 77.6$_{1.1}$ & 81.8$_{0.2}$ & \cg 83.7 \\
\hdashline\noalign{\vskip 0.4ex}
Prompt Tuning  & 78.3$_{2.1}$ & 81.4$_{1.1}$ & 89.3$_{1.4}$   & 74.4$_{0.7}$  & 57.9$_{0.5}$ &  77.8$_{1.6}$ & \cg 76.5 \\
P-Tuning & 82$_{2.2}$ & 82.5$_{0.3}$ & 88.1$_{0.5}$   & 81.9$_{1.7}$  & 67.4$_{0.9}$ & 84.2$_{0.1}$ & \cg 81  \\

SMoP & 80.7$_{1.0}$ & 82.9$_{1.4}$ & 89.8$_{0.3}$ & 78.1$_{2.1}$ & 71.7$_{1.8}$ & 83.7$_{0.9}$ & \cg 81.2 \\

LPT & 81.7$_{0.6}$ & 83.2$_{1.1}$ & 91.8$_{1.3}$   & \textbf{84.3}$_{0.2}$  & 73.6$_{0.7}$ & 84.1$_{0.5}$ & \cg 83.1  \\
\textsc{DePT} & 81.5$_{0.3}$ & \textbf{87.9}$_{1.2}$ & 90.2$_{1.2}$   & 75.7$_{0.6}$  & 71.2$_{1.0}$ & 79.2$_{0.3}$ & \cg 81.0  \\

\method\ (ours)     & \textbf{83.1}$_{0.8}$ & 86.4$_{0.4}$ & \textbf{92.7}$_{1.2}$   & 82.8$_{0.3}$  & \textbf{79.2}$_{0.4}$ & \textbf{84.6}$_{0.5}$ & \cg \textbf{84.8} \\
\midrule
 {\bf Method} & \multicolumn{6}{c}{
\textbf{GLUE (RoBERTa-LARGE Backbone)}} \\                                                       
                 \midrule
Fine-tuning  & 87.6$_{1.7}$ & 94.7$_{2.3}$ & 95.4$_{1.3}$   & 92.1$_{1.2}$  & 88.4$_{0.3}$ & 90.7$_{0.2}$ & \cg 91.48 \\  
\hdashline\noalign{\vskip 0.4ex}
LoRA & 89.1$_{1.1}$ & 87.9$_{0.3}$ & 95.1$_{0.2}$ & 86.5$_{0.9}$ & 78.7$_{0.1}$ & 88.4$_{0.3}$ & \cg 87.6 \\
\hdashline\noalign{\vskip 0.4ex}
Prompt Tuning  & 83.4$_{1.1}$ & 88.2$_{0.2}$ & 92.6$_{0.5}$   & 73.9$_{1.4}$  & 60.8$_{0.6}$ & 81.2$_{0.6}$ & \cg 80.0 \\
P-Tuning & 86.4$_{0.7}$ & 88.7$_{1.2}$ & \textbf{95.8}$_{0.8}$   & 76.3$_{1.1}$  & 62.6$_{0.5}$ & 85.2$_{1.3}$ & \cg 82.5  \\

SMoP & 86.7$_{1.1}$ & 88.4$_{2.2}$ & \textbf{95.8}$_{1.4}$ & 79.6$_{0.8}$ & 76.3$_{1.4}$ & 86.7$_{0.3}$ & \cg 85.6 \\

LPT & 84.2$_{1.1}$ & 86.1$_{0.5}$ & 93.4$_{1.4}$   & \textbf{87.3}$_{0.2}$  & 74.2$_{0.7}$ & 85.3$_{1.3}$ & \cg 85.1  \\
\textsc{DePT} & 83.3$_{1.2}$ & 88.8$_{1.3}$ & 91.2$_{1.8}$   & 77.7$_{0.3}$  & 73.2$_{0.8}$ & 82.2$_{0.7}$ & \cg 82.7  \\

\method\ (ours)     & \textbf{87.4}$_{0.8}$ & \textbf{91.1}$_{0.4}$ & 94.6$_{1.2}$   & 86.1$_{0.3}$  & \textbf{81.1}$_{0.4}$ & \textbf{88.4}$_{0.5}$ & \cg \textbf{88.1}  \\
\midrule
\bottomrule
\end{tabular}
}
\caption{
Test results on GLUE benchmark.
We use RoBERTa-BASE, RoBERTa-LARGE Backbones for all methods. 
We report the score, along with stddev for 3 runs (in the subscript) for all tasks. The best performing Soft Prompt-based method's results are in bold
}
\label{table:main_results_glue_rbase}
\end{table}

We infer that \method\ outperforms all Parameter-Efficient Soft Prompt-based baselines on 4 out of 6 GLUE tasks and w.r.t average task performance, and is a close second for 2 tasks, when using both RoBERTa-BASE and RoBERTa-LARGE backbones. This could be attributed to the attention layer followed by 2-layer MLP in \method, which efficiently generates a context-rich soft prompt. Also, \method\ is shown to be more or similarly efficient compared to well-performing LPT baseline \textit{\textbf{in Section \ref{appendix:model_size_time} of Appendix}}.

\textbf{\textit{Section \ref{appendix:glue_superglue} of Appendix}} shows - \method\ performs better than Soft Prompt baselines - (1) on 2/4 and 3/4 \textbf{SuperGLUE} \cite{wang2019superglue} tasks using RoBERTA-BASE and RoBERTA-LARGE backbones respectively, while giving best average score; (2) when using autoregressive GPT-2 backbone on 3/6 and 2/4 GLUE and SuperGLUE tasks respectively, while giving better average score; (3) on average when using a GPT-2 Large Backbone. 

\noindent \textbf{Comparison with LoRA: }\method\ gives better average score compared to LoRA. 
Specifically, \method\ outperforms LoRA in 5/6 and 3/6 tasks when using RoBERTa-BASE and RoBERTa-LARGE backbones respectively. Also, \method\ is shown to be more efficient than LoRA based on the number of trainable parameters and training and inference times \textit{\textbf{in Section \ref{appendix:model_size_time} of Appendix}}.

\noindent \textbf{Ablation Analysis: }We compare the results of \method\ with just using mean-pooling directly using the RoBERTa-LARGE backbone on 3 GLUE Datasets in Table \ref{tab:ablation}. \method\ outperforms mean-pooling on all 3 tasks, giving an average improvement of $5.82\%$, thus highlighting the importance of the self-attention layer in \method.

\begin{table}[H]
\centering
\resizebox{0.7\columnwidth}{!}{
\begin{tabular}{|l|c|c|c|}
\hline
\textbf{Method}     & \textbf{MRPC} & \textbf{RTE} & \textbf{QQP} \\ \hline
Mean-pooling        & 82.3          & 75.2         & 84.2         \\ \hline
\method             & \textbf{86.1}          & \textbf{81.1}         & \textbf{88.4}         \\ \hline
\end{tabular}
}
\caption{Ablation Analysis on \method}
\label{tab:ablation}
\end{table}

\subsection{Evaluation on SuperGLUE Benchmark}

We compare \method\ with several Soft Prompt-Based Baselines on 4 SuperGLUE Datasets using RoBERTA-BASE and RoBERTA-LARGE backbones in Tables \ref{appendix:glue_superglue} and \ref{tab:superglue-roberta-large} respectively. We observe that \method\ outperforms the baselines on 2/4 and 3/4 tasks using RoBERTA-BASE and RoBERTA-LARGE backbones respectively, while also giving the best average score.

\begin{table}[H]
\centering
\resizebox{\columnwidth}{!}{%
\begin{tabular}{l|cccc
>{\columncolor[HTML]{EFEFEF}}c }
\hline
                & CB   & COPA & MultiRC & BoolQ & Mean \\ \hline
Prompt Tuning   & 75.9 & 52.5 & 67.2    & 63.6  & 64.8 \\
P-Tuning        & 76.3 & 54.7 & 67.9    & 63.7  & 65.6 \\
SMoP            & 79.9 & 57.7 & 67.2    & 69.7  & 68.6 \\
LPT             & 80.6 & \textbf{59.2} & 70.8    & 66.3  & 69.2 \\
\textsc{DePT}            & 78.6 & 52.9 & 67.1    & \textbf{71.4}  & 67.5 \\
\method         & \textbf{83.9} & 57.8 & \textbf{72.9}    & 69.9  & \textbf{71.1} \\ \hline
\end{tabular}%
}
\caption{Test results on 4 SuperGLUE Datasets using RoBERTa-BASE Backbone. The best performing method is bold for each task.}
\label{tab:superglue-roberta}
\end{table}

\begin{table}[H]
\centering
\resizebox{\columnwidth}{!}{%
\begin{tabular}{l|cccc
>{\columncolor[HTML]{EFEFEF}}c }
\hline
                & CB   & COPA & MultiRC & BoolQ & Mean \\ \hline
Prompt Tuning   & 78   & 53   & 67.2    & 63.3  & 65.4 \\
P-Tuning        & 76   & 55   & 68.1    & 64.0  & 65.8 \\
SMoP            & 81.9 & 59   & 69.6    & \textbf{71.1}  & 70.4 \\
LPT             & 82   & \textbf{60}   & 71.0    & 68.0  & 70.2 \\
\textsc{DePT}            & 79   & 54   & 69.0    & 71.0  & 68.2 \\
\method         & \textbf{85}   & \textbf{60}   & \textbf{73.0}    & 70.0  & \textbf{72.0} \\ \hline
\end{tabular}%
}
\caption{Test results on 4 SuperGLUE Datasets using RoBERTa-LARGE Backbone. The best performing method is bold for each task.}
\label{tab:superglue-roberta-large}
\end{table}

\subsection{Zero-Shot Task, Domain Transfer}

Table \ref{tab:domain-transfer-rlarge} shows Zero-Shot Transfer using RoBERTa-LARGE backbone, where a model is trained on training set of a dataset, evaluated on another dataset. We use (QQP, MRPC) and (SST-2, IMDB)\footnote{Task for SST-2 and IMDB is binary classification. SST-2 contains phrases, while IMDB contains full movie reviews} pairs for transfer across tasks and domains respectively similar to \citet{lester2021power}. Table \ref{tab:domain-transfer-rlarge} shows \method\ performs better than Soft Prompt-based baselines, showing \method\ is generalizable across datasets. \method\ even outperforms Fine-tuning in 3/4 pairs. Also, even though \method\ has much less number of parameters compared to LoRA (\textit{\textbf{see Section \ref{appendix:model_size_time} of Appendix}}), \method\ gives better/comparable performance. In addition, we show that \method\ performs better/comparable to well-performing LPT baseline in Few-Shot Task Transfer \textbf{\textit{in Section \ref{appendix:few_shot_transfer} of Appendix.}}

\begin{table}[H]
\footnotesize
\centering
\resizebox{0.95\columnwidth}{!}{
\begin{tabular}{lcccc}
    \toprule
    \textbf{Tuning Method} & \textbf{\begin{tabular}[c]{@{}c@{}}QQP→\\MRPC\end{tabular}} & \textbf{\begin{tabular}[c]{@{}c@{}}MRPC→\\QQP\end{tabular}} & \textbf{\begin{tabular}[c]{@{}c@{}}SST-2→\\IMDB\end{tabular}} & \textbf{\begin{tabular}[c]{@{}c@{}}IMDB→\\SST-2\end{tabular}} \\
    \midrule
    
    Fine-tuning  & 64.0$_{0.7}$  & 68.3$_{1.3}$  & 87.1$_{0.0}$  & 88.8$_{0.4}$ \\ 
    \hdashline\noalign{\vskip 0.4ex}
    LoRA & 71.1$_{0.1}$   & 66.1$_{0.4}$ & 90.3$_{0.2}$ & 87.6$_{1.1}$ \\
    \hdashline\noalign{\vskip 0.4ex}
    Prompt Tuning & 54.1$_{0.3}$  & 54.6$_{0.2}$  & 68.7$_{1.1}$  & 63.5$_{3.8}$ \\
    P-Tuning & 57.6$_{1.2}$  & 52.7$_{1.1}$  & 66.5$_{0.0}$  & 66.8$_{1.3}$ \\
    SMoP & 67.9$_{0.4}$ & 64.1$_{0.6}$ & 84.5$_{0.5}$ & 83.3$_{1.0}$ \\
    LPT & 66.7$_{0.4}$  & 64.5$_{0.3}$  & 67.1$_{0.8}$  & 71.1$_{1.6}$ \\ 
    \textsc{DePT} & 63.3$_{1.8}$  & 58.8$_{0.5}$  & 69.8$_{0.1}$  & 69.3$_{0.9}$ \\      
    \method\ (ours) & \textbf{70.9}$_{1.2}$  & \textbf{69.2}$_{0.4}$  & \textbf{89.1}$_{0.3}$  & \textbf{86.0}$_{0.8}$ \\      
    \bottomrule
\end{tabular}
}
\caption{
    Mean, stddev of zero-shot task, domain transfer for different methods. `Score' is average of Accuracy and macro F1-Score. The best performing Soft Prompt-based method's results are in bold.
}
\label{tab:domain-transfer-rlarge}
\end{table}

\subsection{Method Analysis}\label{sec:method_analysis}

\begin{figure}[H]
    \centering
    \includegraphics[width=0.5\textwidth]{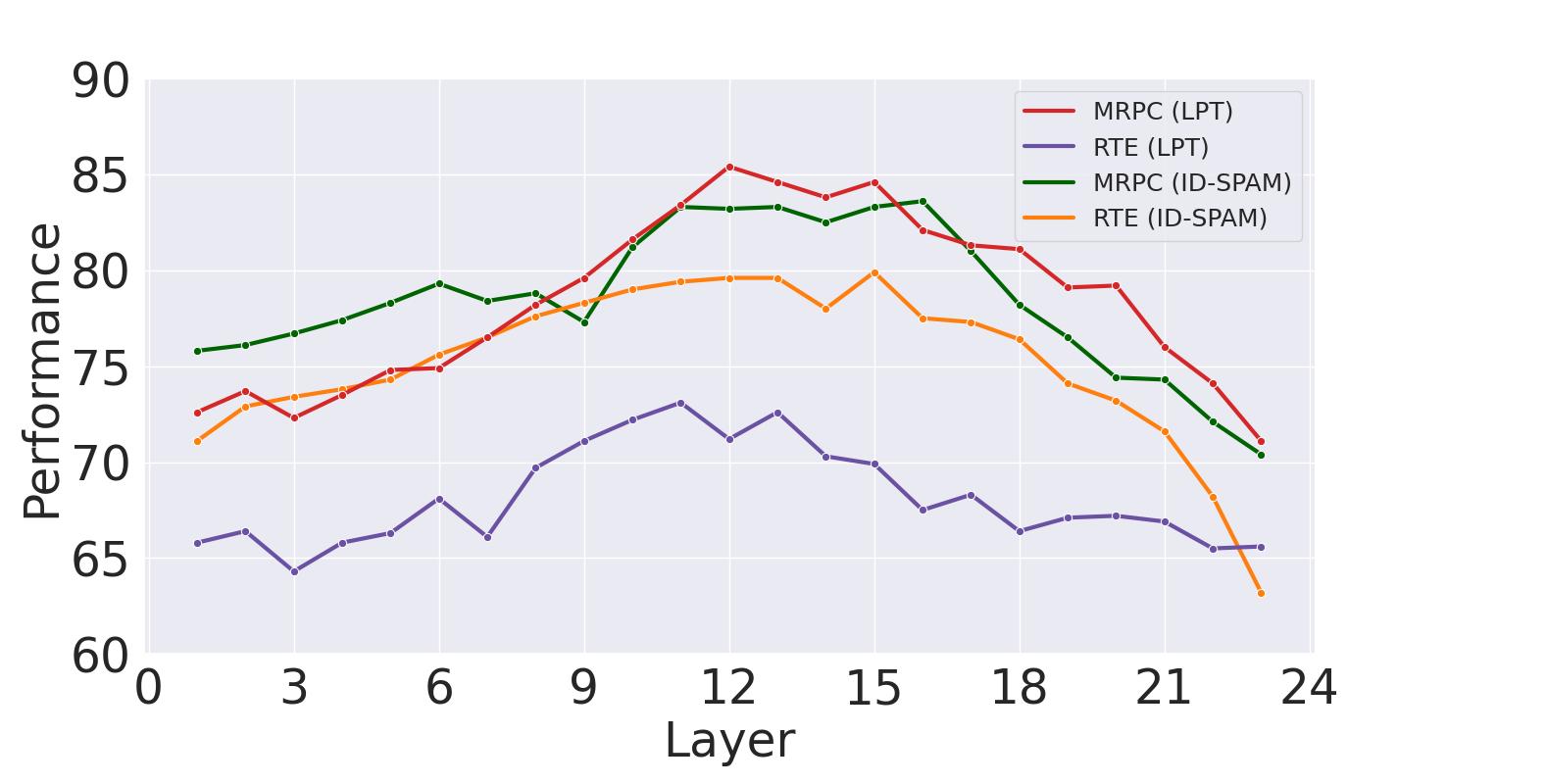}
    \caption{Effect of Variation in layer index ($m$) corresponding to which soft prompt is prepended on performance ($m=0$ refers to input embeddings). Metrics are average of acc. and F1 for MRPC and acc. for RTE. }
    \label{fig:ablation_m}
\end{figure}

We analyze the effect of varying layer index where soft prompt is prepended ($m$ in Figure \ref{fig:id-spam}) on performance of LPT and \method\ on 2 GLUE datasets using RoBERTa-LARGE backbone in Figure \ref{fig:ablation_m}.
We infer that \method\ and LPT perform better when soft prompt is prepended to inputs in middle layers of LM. Also, \method\ significantly outperforms LPT corresponding to almost every layer index for RTE Dataset. Also, \method\ performs better for earlier layers, as soft prompt is generated by using a single attention layer over input embeddings. Hence, prepending this prompt to an earlier layer's outputs performs better than later layer's outputs, as later layer's outputs are obtained after input embeddings are passed through several attention layers, reducing compatibility with the soft prompt. Also, if we prepend soft prompt to later layers, it passes through a small number of layers of LLM, thus showing a reduced performance.

\section{Discussions and Conclusion}\label{sec:disc}

In this paper, we propose \method\ which significantly improves parameter-efficient fine-tuning and zero-shot task and domain transfer performance on various NLU tasks compared to several SOTA parameter-efficient baselines. Notably, further analysis shows that \method\ performs reasonably well when the generated soft prompt is prepended at any layer's inputs. Hence, \method\ is an efficient, input-dependent soft prompt generation framework that could generalize well across several NLP tasks.

\section{Limitations}\label{sec:limit}
We have shown that our proposed approach \method\ improves the performance of two backbone LLMs (RoBERTa-BASE and RoBERTa-LARGE) on multiple NLP tasks. Our framework is generic and can be used with any open source LLMs as backbone. However, we could not use more recent very large scale pre-trained LLMs (like Llama-3.1-70B and Mixtral 8x22B) with tens of billions of parameters as backbone LMs in our experiments due to limited computational resources. We are interested to see the performance gain when we use our approach with those large scale state-of-the-art LLMs in some future work. 

In the current work, we do not have an automated way to choose the layer of the LM where we input the soft prompt. The layer number is kept as a hyperparameter in the current work and its effect is shown in Section \ref{sec:method_analysis}. In future, we want to automatically identify the optimal transformer layer, as proposed by \citet{zhu2023spt}.


\bibliography{SPL}
\newpage
\appendix

\section*{Appendix}


\section{Experiment Settings}
\label{appendix:hyperparams}
For our experiments, we use \texttt{roberta-base} and \texttt{roberta-large} implementations from HuggingFace. For all baselines, the number of appended prompt tokens (for Prompt Tuning, P-tuning, Late Prompt Tuning) are set to 10 tokens. For \textsc{DePT}, we set the rank to 45. For P-Tuning, we set the encoder reparameterization type to MLP. For \method, appended prompt tokens are set to 10 tokens. The search space for hyperparameters for tuning are shown in Table \ref{table:hyperparameters}. For all experiments, standard CrossEntropyLoss was used. For all experiments, we train using a warm-up rate of 0.06, and AdamW optimizer with $\epsilon$ of $1 \times 10^{-6}$, $\beta_1$ of 0.9, $\beta_2$ of 0.98.

In Figure \ref{fig:ablation_m}, we can see that layers 11-13 show optimal performance for both \method\ and LPT. LPT \cite{liu2022late} shows that the 13th layer is optimal. This makes our method \method\ comparable to LPT taking the layer number into account. Also, following the trend from other prior art on soft prompts \cite{lester2021power, liu2022late, li2021prefix, choi2023smop}, we used the best hyperparameter set for each of the baselines. Our experimental approach is also logical and consistent as the experimental settings (choice of backbone LMs, datasets) are same for baselines and our method \method.

\begin{table}[htbp]
\centering
\resizebox{0.5\textwidth}{!}{
\begin{tabular}{|c|c|}
\hline
\textbf{Hyperparameter} & \textbf{Values} \\
\hline
Epochs & \{1, 5, 10, 20, 30\} \\
\hline
Batch Size & \{16, 32, 64\} \\
\hline
Learning Rates & \{1e-3, 5e-4, 1e-4, 5e-3, 1e-5\} \\
\hline
Dropout Rate & \{0.1, 0.2, 0.3\} \\
\hline
Weight Decay & \{0, 0.01, 0.1\} \\
\hline
Layer (RoBERTa-Large) & \{1,2,3...23\} \\
\hline
Layer (RoBERTa-Base) & \{1,2,3...11\} \\
\hline

\end{tabular}
}
\caption{Hyperparameters used for tuning \method.}
\label{table:hyperparameters}
\end{table}

\section{Evaluation using GPT-2 and GPT-2 Large Backbones}
\label{appendix:glue_superglue}



\noindent \textbf{Using GPT-2 Backbone.} We carry out experiments with decoder-only GPT-2 backbone on 6 GLUE Datasets - Table \ref{tab:glue-results-gpt-2} shows that when using GPT-2 as backbone, \method\ outperforms LPT on 3/6 tasks and gives an average performance improvement of $2.3\%$.

\begin{table}[htbp]
\centering
\resizebox{\columnwidth}{!}{%
\begin{tabular}{l|cccccc
>{\columncolor[HTML]{EFEFEF}}c }
\hline
                 & MNLI & QNLI & SST-2 & RTE   & QQP   & MRPC  & AVG \\ \hline
LPT              & 69.5 & \textbf{79.4} & \textbf{90.1}  & 62.8  & 80.3  & \textbf{81.9}  & 77.3 \\
\method          & \textbf{78.3} & 77.1 & 85.1  & \textbf{71.6}  & \textbf{82.9}  & 79.5  & \textbf{79.1} \\ \hline
\end{tabular}%
}
\caption{Test results on 6 GLUE Datasets using GPT-2 Backbone. The best performing PEFT method is bold for each task.}
\label{tab:glue-results-gpt-2}
\end{table}

 Next, we carry out experiments with decoder-only GPT-2 backbone on 4 SuperGLUE Datasets –
Table \ref{tab:superglue-gpt-2} shows that compared to Soft Prompt-Based baselines, \method\ gives the best average score, and performs the best on 2 tasks, while performing the second best on one of them.

\begin{table}[H]
\centering
\resizebox{\columnwidth}{!}{%
\begin{tabular}{l|cccc
>{\columncolor[HTML]{EFEFEF}}c }
\hline
                 & CB   & COPA & MultiRC & BoolQ & Mean \\ \hline
Prompt Tuning    & 71.7 & 57   & 61.7    & 64.1  & 63.6 \\
P-Tuning         & 73.3 & 57.7 & 63.2    & 65.7  & 65   \\
SMoP             & 81.4 & 61.2 & 68.4    & 69.4  & 70.1 \\
LPT              & 82.1 & 61.3 & 72.1    & \textbf{74.1}  & 72.4 \\
\textsc{DePT}             & 76.1 & 55.1 & \textbf{73.5}    & 67.2  & 68   \\
\method          & \textbf{88.1} & \textbf{63.1} & 71.7    & 72.4  & \textbf{73.8} \\ \hline
\end{tabular}%
}
\caption{Test results on 4 SuperGLUE Datasets using GPT-2 Backbone. The best performing method is bold for each task.}
\label{tab:superglue-gpt-2}
\end{table}

\noindent \textbf{Using GPT-2 Large Backbone. } We compare the performance of \method\ with LoRA and LPT using a large generative model GPT-2 Large (around 0.8 Billion Parameters) as the backbone on 2 GLUE Datasets - RTE and MRPC, as shown in Table \ref{tab:gpt2-large-glue}.

\begin{table}[H]
    \centering
    \begin{tabular}{lccc}
        \toprule
        Method & RTE & MRPC & Average \\
        \midrule
        LoRA & 74.0 & 80.0 & 77.0 \\
        LPT & 69.9 & 82.9 & 76.4 \\
        \method & 73.7 & 81.1 & \textbf{77.4} \\
        \bottomrule
    \end{tabular}
    \caption{Test results on 2 GLUE Datasets using GPT-2 Large Backbone.}
    \label{tab:gpt2-large-glue}
\end{table}

\method\ gives an average improvement of 0.5\% and 1.3\% compared to LoRA and LPT respectively across the 2 GLUE Datasets, showing that \method\ is competitive even for a large, generative backbone LM.

\section{Few-Shot Task Transfer}
\label{appendix:few_shot_transfer}


\begin{table}[H]
    \centering
    \resizebox{\columnwidth}{!}{%
        \begin{tabular}{|l|l|l|c|}
            \hline
            \textbf{Train} & \textbf{Eval (Few-shot, 100 samples)} & \textbf{Tuning} & \textbf{Score} \\ \hline
            MRPC & QQP & Fine-Tuning & 81.7 \\ 
            MRPC & QQP & LPT & 74.4 \\ 
            MRPC & QQP & \method & 73.1 \\ \hline
            QQP & MRPC & Fine-Tuning & 79.7 \\ 
            QQP & MRPC & LPT & 69.4 \\ 
            QQP & MRPC & \method & 72.5 \\ \hline
        \end{tabular}%
    }
    \caption{Few-shot task transfer for different methods using the RoBERTa-LARGE Backbone.}
    \label{tab:few-shot-transfer}
\end{table}

\method\ and LPT (a well-performing baseline in Table \ref{table:main_results_glue_rbase}) using the RoBERTa-LARGE Backbone are fine-tuned on the first dataset, and then further fine-tuned on 100 training samples from the second. This model is then evaluated on the second dataset.

From Table \ref{tab:few-shot-transfer}, we can see that \method\ performs better than LPT on QQP->MRPC, while the performance is comparable for MRPC->QQP.

\section{Comparison of \method\ with baselines w.r.t model size and training and inference times}
\label{appendix:model_size_time}


\begin{table}[H]
    \centering
    \resizebox{\columnwidth}{!}{%
        \begin{tabular}{|l|c|c|c|}
            \hline
            \textbf{Model} & \textbf{LPT} & \textbf{LoRA} & \method \\ \hline
            RoBERTa-BASE & 2,162,688 & 3,495,312 & 2,064,384 \\ \hline
            RoBERTa-LARGE & 2,883,584 & 7,931,280 & 3,538,944 \\ \hline
        \end{tabular}%
    }
    \caption{number of trainable parameters of \method\ and well-performing baselines LPT and LoRA (see Table \ref{table:main_results_glue_rbase}).}
    \label{tab:parameters}
\end{table}

Table \ref{tab:parameters} shows that the number of trainable parameters in \method\ is lesser than that of LoRA for both backbones, and is lesser than that of LPT using RoBERTa-BASE backbone, while they are comparable in case of RoBERTa-LARGE backbone.

\begin{table}[H]
\centering
\resizebox{\columnwidth}{!}{%
\begin{tabular}{|l|c|c|c|}
\hline
\textbf{Backbone} & \textbf{\begin{tabular}[c]{@{}c@{}}No. of Parameters\\in Backbone LM\end{tabular}} & \textbf{\method} & \textbf{LoRA} \\ \hline
GPT2 & 126.8 & 2.1 & 2.4 (1.1x) \\ \hline
GPT2-medium & 361.1 & 3.5 & 6.3 (1.8x) \\ \hline
GPT2-large & 785.8 & 5.1 & 11.8 (2.3x) \\ \hline
GPT2-xl & 1577.3 & 8.3 & 19.7 (2.4x) \\ \hline
Gemma-2B \cite{team2024gemma} & 2525.8 & 13.4 & 19.6 (1.5x) \\ \hline
FLAN-T5-xl \cite{flan} & 2823.6 & 13.4 & 35.5 (2.6x) \\ \hline
\end{tabular}%
}
\caption{Number of trainable parameters (in millions) of \method\ compared to LoRA for several LM backbones of different sizes. The decrease in the number of trainable parameters of \method\ compared to LoRA is written within brackets.}
\label{tab:large-sizes-params}
\end{table}

Table \ref{tab:large-sizes-params} shows that as the size of the backbone LM increases, efficiency in the number of trainable parameters of \method\ compared to LoRA tends to increase. Hence, \method\ is suitable even for massive LMs.

\begin{table}[H]
    \centering
    \resizebox{\columnwidth}{!}{%
        \begin{tabular}{|l|l|c|c|}
            \hline
            \textbf{Dataset} & \textbf{Method} & \textbf{\begin{tabular}[c]{@{}c@{}}Training Time per\\ sample (in secs)\end{tabular}} & \textbf{\begin{tabular}[c]{@{}c@{}}Inference Time per\\ sample (in secs)\end{tabular}} \\ \hline
            BoolQ & LPT & 0.669 & 0.236 \\ \hline
            BoolQ & LoRA & 0.715 & 0.313 \\ \hline
            BoolQ & \method & 0.651 & 0.251 \\ \hline
            WiC & LPT & 0.082 & 0.041 \\ \hline
            WiC & LoRA & 0.113 & 0.067 \\ \hline
            WiC & \method & 0.084 & 0.035 \\ \hline
        \end{tabular}%
    }
    \caption{Training and inference times of ID-SPAM and well-performing baselines LPT and LoRA for 2 SuperGLUE Datasets.}
    \label{tab:times}
\end{table}

Table \ref{tab:times} shows that \method\ requires less time for training as well as for inference, in comparison to LoRA on both BoolQ (a yes/no QA dataset) and WiC (dataset for binary classification) Datasets (2 datasets from SuperGLUE). Also, \method\ takes lesser time to train on BoolQ than LPT, while the times are comparable on WiC. In case of inference, \method\ takes lesser time than LPT for WiC, while taking slightly more time than LPT for BoolQ. Hence, \method\ has comparable training and inference times w.r.t LPT, while giving better performance on GLUE datasets (see Table \ref{table:main_results_glue_rbase}).

\begin{table}[H]
\centering
\resizebox{\columnwidth}{!}{%
\begin{tabular}{lcccccc}
\toprule
 & MNLI & QNLI & SST-2 & RTE & QQP & MRPC \\
\midrule
Fine Tuning & 2887s & 270s & 224s & 247s & 1854s & 87s \\
LPT & 2013s & 157s & 161s & 168s & 1157s & 59s \\
\method & 1902s & 166s & 171s & 168s & 1004s & 51s \\
\bottomrule
\end{tabular}
}
\caption{Total training time cost before convergence (in seconds) of \method\ compared to baselines}
\label{tab:convergence_times}
\end{table}

Table \ref{tab:convergence_times} shows the training convergence times (in seconds - lower the better) for LPT and our proposed \method\ (By convergence, we mean the epoch where the validation error is the least) using RoBERTa-LARGE Backbone. We can see that \method\ gives better/similar convergence time compared to LPT on 4 out of 6 GLUE Tasks. Also, LPT takes an average convergence of time of 619 s, while ID-SPAM takes 577 s, giving an improvement of $7.3\%$ in average convergence time.

\section{Convergence of the LoRA Baseline}
\label{appendix:lora-convergence}

The training loss is tabulated every 5 epochs in Table \ref{tab:lora-loss} when training LoRA with the RoBERTa-BASE backbone on the MRPC and RTE Datasets from the GLUE Benchmark.

\begin{table}[H]
\centering
\begin{tabular}{|c|c|c|}
\hline
\textbf{Epoch} & \textbf{MRPC} & \textbf{RTE} \\ \hline
5              & 0.21          & 0.4          \\ \hline
10             & 0.12          & 0.14         \\ \hline
15             & 0.05          & 0.07         \\ \hline
20             & 0.02          & 0.06         \\ \hline
25             & 0.02          & 0.04         \\ \hline
30             & 0.0001        & 0.02         \\ \hline
\end{tabular}
\caption{Training Loss across epochs when training LoRA with the RoBERTa-BASE backbone}
\label{tab:lora-loss}
\end{table}

We can see that the training loss continuously decreases with increasing epochs on both the MRPC and RTE Datasets. Also, the losses are considerably lowered after 30 epochs as can be seen in the table, thus showing convergence.

\end{document}